\documentclass[11pt,a4paper]{article}
\usepackage[hyperref]{acl2017}
\usepackage{times}
\usepackage{latexsym}
\usepackage{url}
\usepackage{graphicx}
\usepackage{subcaption}
\usepackage{amssymb}
\usepackage{amsfonts}
\usepackage{amsmath}
\usepackage{graphicx}
\usepackage{xspace}
\usepackage{xcolor}
\usepackage{pifont}
\usepackage{enumitem}
\usepackage{url}
\usepackage{latexsym}
\usepackage{array}
\usepackage{multirow}
\usepackage{arydshln}
\graphicspath{ {./}}

\aclfinalcopy

\newcommand{\pos}{\textsc{pos}\xspace}
\newcommand{\crf}{\textsc{crf}\xspace}
\newcommand{\rnn}{\textsc{rnn}\xspace}
\newcommand{\arnn}{\textit{a}-\textsc{rnn}\xspace}
\newcommand{\darnn}{\textit{da}-\textsc{rnn}\xspace}

\newcommand{\danornn}{\textit{da}-\textsc{cent}\xspace}
\newcommand{\nornn}{\textit{eq}-\textsc{cent}\xspace}
\newcommand{\eqrnn}{\textit{eq}-\textsc{rnn}\xspace}
\newcommand{\cnn}{\textsc{cnn}\xspace}

\newcommand{\bow}{\textsc{bow}\xspace}
\newcommand{\mlp}{\textsc{mlp}\xspace}
\newcommand{\lr}{\textsc{lr}\xspace}
\newcommand{\gbdt}{\textsc{gbdt}\xspace}
\newcommand{\WordVec}{\textsc{word2vec}\xspace}
\newcommand{\CBOW}{\textsc{cbow}\xspace}

\newcommand{\GazzettaTrain}{\textsc{g-train-l}\xspace}
\newcommand{\GazzettaTrainS}{\textsc{g-train-s}\xspace}
\newcommand{\GazzettaDev}{\textsc{g-dev}\xspace}
\newcommand{\GazzettaTest}{\textsc{g-test-l}\xspace}
\newcommand{\GazzettaClean}{\textsc{g-test-s}\xspace}
\newcommand{\GazzettaCleanR}{\textsc{g-test-s-r}\xspace}
\newcommand{\DetoxAttTrain}{\textsc{w-att-train}\xspace}
\newcommand{\DetoxAttDev}{\textsc{w-att-dev}\xspace}
\newcommand{\DetoxAttTest}{\textsc{w-att-test}\xspace}

\newcommand{\DetoxToxTrain}{\textsc{w-tox-train}\xspace}
\newcommand{\DetoxToxDev}{\textsc{w-tox-dev}\xspace}
\newcommand{\DetoxToxTest}{\textsc{w-tox-test}\xspace}
\newcommand{\R}{\mathbb{R}}
\newcommand{\Detox}{\textsc{detox}\xspace}
\newcommand{\gru}{\textsc{gru}\xspace}
\newcommand{\lstm}{\textsc{lstm}\xspace}
\newcommand{\auc}{\textsc{auc}\xspace}
\newcommand{\relu}{\textsc{r}e\textsc{lu}\xspace}
\newcommand{\glove}{\textsc{glove}\xspace}
\newcommand{\oov}{\textsc{oov}\xspace}
\newcommand{\modlist}{\textsc{list}\xspace}

\newcommand{\svm}{\textsc{svm}\xspace}
\title{Deep Learning for User Comment Moderation}

\author{
    John Pavlopoulos \\
    StrainTek\\Athens, Greece \\
    {\tt ip@straintek.com} \\\And
    Prodromos Malakasiotis \\
    StrainTek\\Athens, Greece \\ 
    {\tt mm@straintek.com} \\\And
    Ion Androutsopoulos \\
    Department of Informatics \\
    Athens University of Economics \\and Business, Greece \\
   {\tt ion@aueb.gr} \\
   }
  
\date{}

\begin{document}
\maketitle
\begin{abstract}
Experimenting with a new dataset of 1.6M user comments from a Greek news portal and existing datasets of English Wikipedia comments, we show that an \rnn outperforms the previous state of the art in moderation. A deep, classification-specific attention mechanism improves further the overall performance of the \rnn. We also compare against a \cnn and a word-list baseline, considering both fully automatic and semi-automatic moderation.  
\end{abstract}


\section{Introduction}

User comments play a central role in social media and online discussion fora. News portals and blogs often also allow their readers to comment in order to get feedback, engage their readers, and build customer loyalty.
User comments, however, and more generally user content can also be abusive (e.g., bullying, profanity, hate speech). Social media are increasingly under pressure to combat abusive content. News portals also suffer from abusive user comments, which damage their reputation and make them liable to fines, e.g., when hosting comments encouraging illegal actions. They often employ moderators, who are frequently overwhelmed by the volume of comments. Readers are disappointed when non-abusive comments do not appear quickly online because of moderation delays. Smaller news portals may be unable to employ moderators, and some are forced to shut down their comments sections entirely.\footnote{See, for example,  \url{http://niemanreports.org/articles/the-future-of-comments/}.} 

We examine how deep learning \cite{Goodfellow2016,Goldberg2016} can be used to moderate user comments. We experiment with a new dataset of approx.\ 1.6M manually moderated user comments from a Greek sports portal (Gazzetta), which we make publicly available.\footnote{The portal is \url{http://www.gazzetta.gr/}. Instructions to obtain the Gazzetta data will be posted at \url{http://nlp.cs.aueb.gr/software.html}.} Furthermore, we provide word embeddings pre-trained on 5.2M comments from the same portal. We also experiment on the datasets of Wulczyn et al.\ \shortcite{Wulczyn2016}, which contain English Wikipedia  comments labeled for personal attacks,  aggression, toxicity. 

\begin{figure}
	\centering
	\includegraphics[width=0.75\columnwidth]{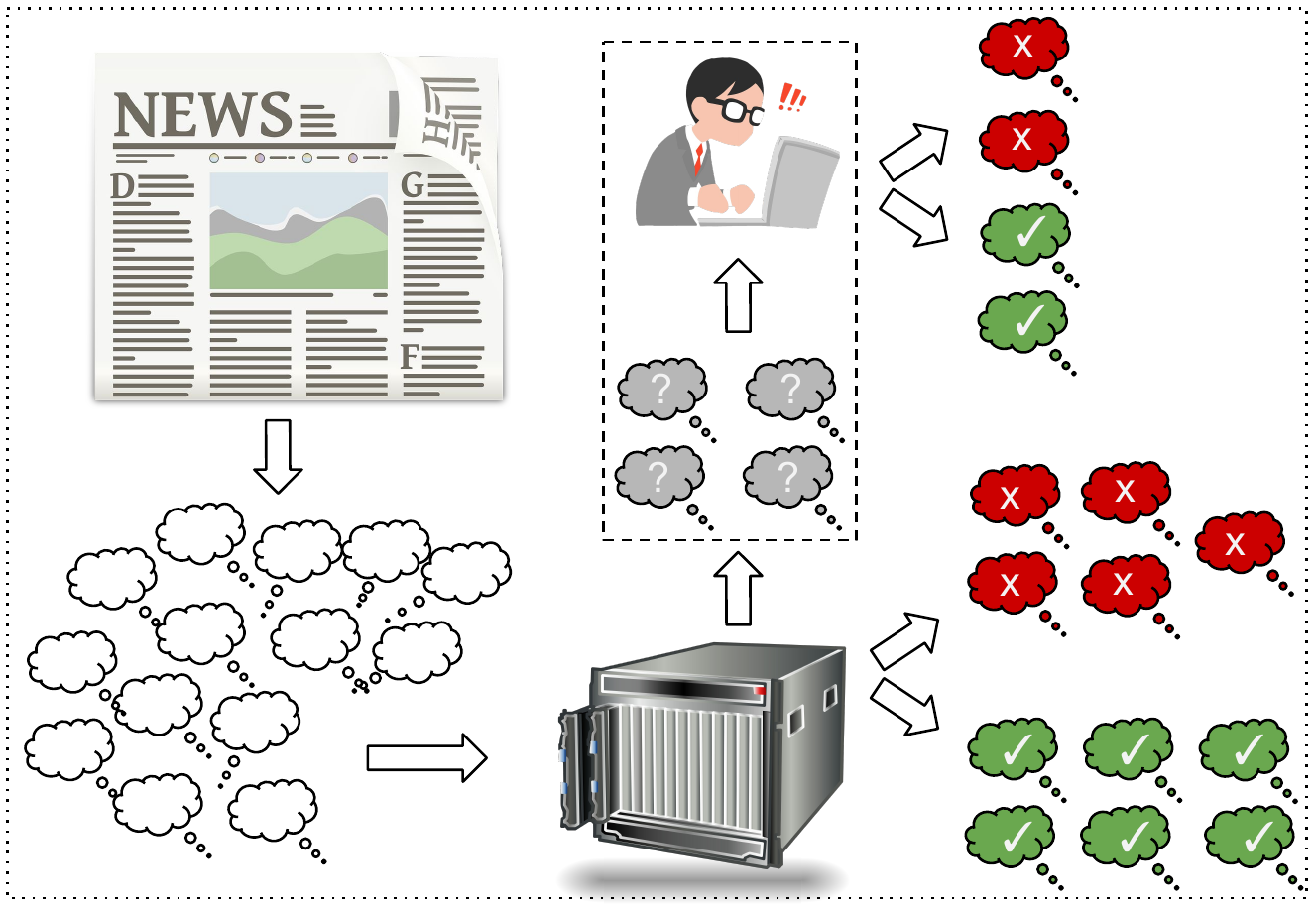}
	\caption{Semi-automatic moderation.}
	\label{fig:architecture}
\end{figure}

In a fully automatic scenario, a system directly accepts or rejects comments. Although this scenario may be the only available one, e.g., when portals cannot afford moderators, it is unrealistic to expect that fully automatic moderation will be perfect, because abusive comments may involve irony, sarcasm, harassment without profanity etc., which are particularly difficult for machines to handle. When moderators are available, it is more realistic to develop semi-automatic systems to assist rather than replace them, a scenario that has not been considered in previous work. Comments for which the system is uncertain (Fig.~\ref{fig:architecture}) are shown to a moderator to decide; all other comments are accepted or rejected by the system. We discuss how moderation systems can be tuned, depending on the availability and workload of moderators. We also introduce additional evaluation measures for the semi-automatic scenario.

On both Gazzetta and Wikipedia comments and for both scenarios (automatic, semi-automatic), we show that a recursive neural network (\rnn) outperforms the system of Wulczyn et al.\ \shortcite{Wulczyn2016}, the previous state of the art for comment moderation, which employed logistic regression (\lr) or a multi-layered Perceptron (\mlp). We also propose an attention mechanism that improves the overall performance of the \rnn. Our attention differs from most previous ones \cite{Bahdanau2015,Luong2015} in that it is used in text classification, where there is no previously generated output subsequence to drive the attention, unlike sequence-to-sequence models \cite{Sutskever2014}. In effect, our attention mechanism detects the words of a comment that affect mostly the classification decision (accept, reject), by examining them in the context of the particular comment. 

Our main contributions are: (i) We release a new dataset of 1.6M moderated user comments. (ii) We are among the first to apply deep learning to user comment moderation, and we show that an \rnn with a novel classification-specific attention mechanism outperforms the previous state of the art. (iii) Unlike previous work, we also consider a semi-automatic scenario, along with threshold tuning and evaluation measures for it. 


\section{Datasets}

We first discuss the datasets we used, to help acquaint the reader with the problem. 

\begin{table}
\centering
{\small
\begin{tabular}{|c|c|c|c|}
\hline
Dataset/Split   & Accepted       &   Rejected     &  Total  \\\hline
\GazzettaTrain  & 960,378 (66\%) & 489,222 (34\%) & 1,45M   \\
\GazzettaTrainS & 67,828 (68\%)  & 32,172 (32\%)  & 100,000 \\
\GazzettaDev    & 20,236 (68\%)  & 9,464 (32\%)   & 29,700  \\
\GazzettaTest   & 20,064 (68\%)  & 9,636 (32\%)   & 29,700  \\
\GazzettaClean  & 1,068 (71\%)   & 432 (29\%)     & 1,500   \\
\GazzettaCleanR & 1,174 (78\%)   & 326 (22\%)     & 1,500   \\\hline
\DetoxAttTrain  & 61,447 (88\%)  & 8,079 (12\%)   & 69,526  \\ 
\DetoxAttDev    & 20,405 (88\%)  & 2,755 (12\%)   & 23,160  \\ 
\DetoxAttTest   & 20,422 (88\%)  & 2,756 (12\%)   & 23,178  \\\hline
\DetoxToxTrain  & 86,447 (90\%)  & 9,245 (10\%)   & 95,692  \\ 
\DetoxToxDev    & 29,059 (90\%)  & 3,069 (10\%)   & 32,128  \\ 
\DetoxToxTest   & 28,818 (90\%)  & 3,048 (10\%)   & 31,866  \\\hline
\end{tabular}
}
\caption{Statistics of the datasets used.}
\label{tab:data_stats}
\end{table}


\subsection{Gazzetta dataset} \label{sec:Gazzetta}

There are approx.\ 1.45M training comments (covering Jan.\ 1, 2015 to Oct.\ 6, 2016) in the Gazzetta dataset; we call  them \GazzettaTrain (Table~\ref{tab:data_stats}). Some experiments use only the first 100K comments of \GazzettaTrain, called \GazzettaTrainS. An additional set of 60,900 comments (Oct.\ 7 to Nov.\ 11, 2016) was split to development (\GazzettaDev, 29,700 comments), large test (\GazzettaTest, 29,700), and small test set (\GazzettaClean, 1,500). Gazzetta's moderators (2 full-time, plus journalists occasionally helping) are occasionally instructed to be stricter (e.g., during violent events). To get a more accurate view of performance in normal situations, we manually re-moderated (labeled as `accept' or `reject') the comments of \GazzettaClean, producing \GazzettaCleanR. The reject ratio is approximately 30\% in all subsets, except for \GazzettaCleanR where it drops to 22\%, because there are no occasions where the moderators were instructed to be stricter in \GazzettaCleanR.

Each \GazzettaCleanR comment was re-moderated by 5 annotators. Krippendorff's \shortcite{Krippendorff2004} alpha was 0.4762, close to the value (0.45) reported by Wulczyn et al.\ \shortcite{Wulczyn2016} for Wikipedia comments. Using Cohen's Kappa \cite{Cohen1960}, the mean pairwise agreement was 0.4749. The mean pairwise percentage of agreement (\% of comments each pair of annotators agreed on) was 81.33\%. Cohen's Kappa and Krippendorff's alpha lead to moderate scores, because they account for agreement by chance, which is high when there is class imbalance (22\% reject, 78\% accept in \GazzettaCleanR). 

We also provide 300-dimensional word embeddings, pre-trained on approx.\ 5.2M comments (268M tokens) from Gazzetta using \WordVec \cite{Mikolov2013a, Mikolov2013c}.\footnote{We used \CBOW, window size 5, min.\ term freq.\ 5, negative sampling, obtaining a vocabulary size of approx.\ 478K.} This larger dataset cannot be used to train classifiers, because most of its comments are from a period (before 2015) when Gazzetta did not employ moderators.


\subsection{Wikipedia datasets} \label{sec:wikipedia}

Wulczyn et al.~\shortcite{Wulczyn2016} created three datasets containing English Wikipedia talk page comments.

\smallskip
\noindent\textbf{Attacks dataset:} This dataset contains approx.\ 115K comments, which were labeled as personal attacks (reject) or not (accept) using crowdsourcing. Each comment was labeled by at least 10 annotators. Inter-annotator agreement, measured on a random sample of 1K comments using Krippendorff's \shortcite{Krippendorff2004} alpha, was 0.45. The gold label of each comment is determined by the majority of annotators, leading to \emph{binary labels} (accept, reject). Alternatively, the gold label is the percentage of annotators that labeled the comment as `accept' (or `reject'), leading to \emph{probabilistic labels}.\footnote{We also construct probabilistic gold labels (in addition to binary ones) for \GazzettaCleanR, where there are 5 annotators.} The dataset is split in three parts (Table~\ref{tab:data_stats}): training (\DetoxAttTrain, 69,526 comments), development (\DetoxAttDev, 23,160), and test (\DetoxAttTest,  23,178 comments). In all three parts, the rejected comments are 12\%, but this ratio is artificial (in effect, Wulczyn et al.\ oversampled comments posted by banned users), unlike Gazzetta subsets where the truly observed accept/reject ratios are used.

\smallskip
\noindent\textbf{Toxicity dataset:} This dataset was created like the previous one, but contains more comments (159,686), now labeled as toxic (reject) or not (accept). Inter-annotator agreement was not reported. Again, binary or probabilistic gold labels can be used. The dataset is split in three parts (Table~\ref{tab:data_stats}): training (\DetoxToxTrain, 95,692 comments), development (\DetoxToxDev, 32,128), and test (\DetoxToxTest, 31,866). In all three parts, the rejected (toxic) comments are 10\%, again an artificial ratio.

Wikipedia comments are longer (median 38 and 39 tokens for attacks, toxicity) compared to Gazzetta's (median 25).  Wulczyn et al.\ \shortcite{Wulczyn2016} also created an `aggression' dataset containing the same comments as the personal attacks one, but now labeled as aggressive or not.  The (probabilistic) labels of the two datasets are very highly correlated (0.8992 Spearman, 0.9718 Pearson) and we do not consider the aggression dataset further. 


\section{Methods}

We experimented with an \rnn operating on word embeddings, the same \rnn enhanced with our attention mechanism (\arnn), several variants of \arnn, a vanilla convolutional neural network (\cnn) also operating on word embeddings, the \Detox system of Wulczyn et al.\ \shortcite{Wulczyn2016}, and a baseline that uses word lists with precision scores.


\subsection{DETOX} \label{sec:detox}

\Detox \cite{Wulczyn2016} was the previous state of the art in comment moderation, in the sense that it had the best reported results on the Wikipedia datasets (Section~\ref{sec:wikipedia}), the largest previous publicly available datasets of moderated user comments.\footnote{Two of the co-authors of Wulczyn et al.\ \shortcite{Wulczyn2016} are with Jigsaw, who recently announced Perspective, a system to detect `toxic' comments. Perspective is not the same as \Detox (personal communication), but we were unable to obtain scientific articles describing it. We have applied for access to its \textsc{api} (\url{http://www.perspectiveapi.com/}).} \Detox represents each comment as a bag of word $n$-grams ($n \leq 2$, each comment becomes a bag containing its 1-grams and 2-grams) or a bag of character $n$-grams ($n \leq 5$, each comment becomes a bag containing character 1-grams, \dots, 5-grams). \Detox can rely on a logistic regression (\lr) or multi-layer Perceptron (\mlp) classifier, and uses binary or probabilistic gold labels (Section~\ref{sec:wikipedia}) during training. We used the \Detox implementation of Wulczyn et al.\ and the same grid search to tune the hyper-parameters that select word or character $n$-grams, classifier (\lr or \mlp), and gold labels (binary or probabilistic). For Gazzetta, only binary gold labels were possible, since \GazzettaTrain and \GazzettaTrainS have a single gold label per comment. Unlike Wulczyn et al., we tuned the hyper-parameters by evaluating (computing \auc and Spearman, Section~\ref{sec:results}) on a random 2\% of held-out comments of \DetoxAttTrain, 
\DetoxToxTrain, or \GazzettaTrainS, instead of the development subsets, to be able to obtain more realistic results from the development sets while developing the methods. The tuning always preferred character $n$-grams, as in the work of Wulczyn et al., and \lr to \mlp, whereas Wulczyn et al.\ reported slightly higher performance for the \mlp on \DetoxAttDev.\footnote{Wulczyn et al.\ \shortcite{Wulczyn2016} report results only on \DetoxAttDev. We repeated the tuning by evaluating on \DetoxAttDev, and again character $n$-grams with \lr were selected.} The tuning also selected probabilistic labels when available (Wikipedia datasets), as in the work of Wulczyn et al. 


\subsection{RNN-based methods}

\textbf{\rnn:} The \rnn method is a chain of \gru cells \cite{Cho2014} that transforms the tokens $w_{1} \dots, w_{k}$ of each comment to hidden states $h_{1} \dots, h_{k}$, followed by an \lr layer that uses $h_k$ to classify the comment (accept, reject). Formally, given the vocabulary $V$, a matrix $E \in \R^{d \times |V|}$ containing $d$-dimensional word embeddings, an initial $h_0$, and a comment $c = \left<w_{1}, \dots, w_{k}\right>$, the \rnn computes $h_{1}, \dots, h_{k}$ as follows ($h_t \in \R^m$):
\begin{eqnarray*}
\tilde{h}_{t} &=& \tanh (W_{h} x_{t} + U_{h} (r_{t} \odot h_{t-1}) + b_{h}) \\
h_{t} &=& (1 - z_{t}) \odot h_{t-1} + z_{t} \odot \tilde{h}_{t}\\
z_{t} &=& \sigma(W_{z} x_{t} + U_{z} h_{t-1} + b_{z})\\
r_{t} &=& \sigma(W_{r} x_{t} + U_{r} h_{t-1} + b_{r})
\end{eqnarray*}
where $\tilde{h}_{t} \in \R^m$ is the proposed hidden state at position $t$, obtained by considering the word embedding $x_t$ of token $w_t$ and the previous hidden state $h_{t-1}$; $\odot$ denotes element-wise multiplication; $r_t \in \R^m$ is the reset gate (for $r_t$ all zeros, it allows the \rnn to forget the previous state $h_{t-1}$); $z_t \in \R^m$ is the update gate (for $z_t$ all zeros, it allows the \rnn to ignore the new proposed $\tilde{h}_{t}$, hence also $x_t$, and copy $h_{t-1}$ as $h_t$); $\sigma$ is the sigmoid function; $W_h, W_z, W_r  \in \R^{m \times d}$; $U_h, U_z, U_r \in \R^{m \times m}$; $b_h, b_z, b_r \in \R^m$. Once $h_k$ has been computed, the \lr layer estimates the  probability that comment $c$ should be rejected, with  $W_p \in \R^{1 \times m}, b_p \in \R$:
\begin{eqnarray*}
P_{\rnn}(\textit{reject}|c) &=& \sigma(W_{p} h_{k} + b_{p}) 
\end{eqnarray*}

\begin{figure}
	\centering
	\includegraphics[width=\columnwidth]{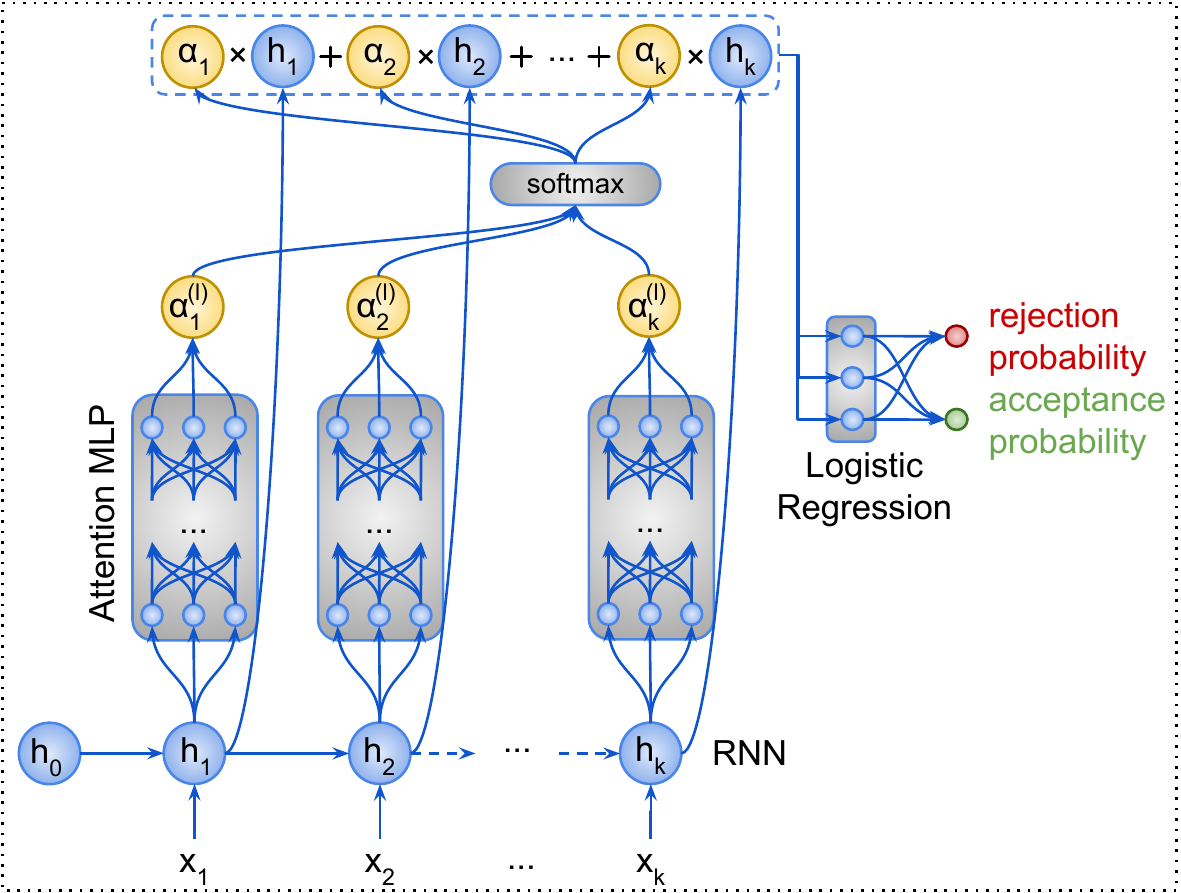}
	\caption{Illustration of \arnn.}
	\label{fig:arnn}
\end{figure}

\noindent \textbf{\arnn:} When the attention mechanism is added, the \lr layer considers the weighted sum $h_{\textit{sum}}$ of all the hidden states, instead of just $h_k$ (Fig.~\ref{fig:arnn}):
\vspace*{-2mm}
\begin{eqnarray}
h_{sum} &=& \sum_{t=1}^{k} a_{t} h_{t} \label{eq:hsum} \\
P_{\arnn}(\textit{reject}|c) &=& \sigma(W_{p} h_{\textit{sum}} + b_{p}) \nonumber
\end{eqnarray}
The weights $a_t$ are produced by an attention mechanism, which is an \mlp with $l$ layers:
\begin{eqnarray}
a_t^{(1)} &=& \relu(W^{(1)} h_{t} + b^{(1)}) \label{eq:arnn} \\
& \dots & \nonumber \\
a_t^{(l-1)} &=& \relu(W^{(l-1)} a_t^{(l-2)} + b^{(l-1)}) \nonumber \\
a_t^{(l)} &=& W^{(l)} a_t^{(l-1)} + b^{(l)} \nonumber \\
a_t &=& \texttt{softmax}(a_t^{(l)}; a_1^{(l)}, \dots, a_k^{(l)}) \nonumber 
\end{eqnarray}
where $a_t^{(1)}, \dots, a_t^{(l-1)} \in \R^{r}$, $a_t^{(l)}, a_t \in \R$, 
$W^{(1)} \in \R^{r \times m}$, $W^{(2)}, \dots, W^{(l-1)} \in \R^{r \times r}$, $W^{(l)} \in \R^{1 \times r}$,
$b^{(1)}, \dots, b^{(l-1)} \in \R^r$, $b^{(l)} \in \R$. 
The \texttt{softmax} operates across all the $a_t^{(l)}$ ($t=1, \dots, k$), making the attention weights $a_t$ sum to 1.
Our attention mechanism differs from most previous ones \cite{Mnih2014,Bahdanau2015,Xu2015,Luong2015} in that it is used in a classification setting, where there is no previously generated output subsequence (e.g., partly generated translation) to drive the attention (e.g., assign more weight to source words to translate next), unlike seq2seq models \cite{Sutskever2014}. It  assigns larger weights $a_t$ to hidden states $h_t$ corresponding to positions where there is more evidence that the comment should be accepted or rejected. 

Yang et al.\ \shortcite{Yang2016} use a similar attention mechanism, but ours is deeper. In effect they always set $l=2$, whereas we allow $l$ to be larger (tuning selects $l=4$).\footnote{Yang et al.\ use $\tanh$ instead of \relu in Eq.~\ref{eq:arnn}, which works worse in our case, and no bias $b^{(l)}$ in the $l$-th layer.} On the other hand, the attention mechanism of Yang et al.\ is part of a classification method for longer texts (e.g., product reviews). Their method uses two \gru \rnn{s}, both bidirectional \cite{Schuster1997}, one turning the word embeddings of each sentence to a sentence embedding, and one turning the sentence embeddings to a document embedding, which is then fed to an \lr layer. Yang et al.\ use their attention mechanism in both \rnn{s}, to assign attention scores to words and sentences. We consider shorter texts (comments), we have a single \rnn, and we assign attention scores to words only.\footnote{We tried a bidirectional instead of unidirectional \gru chain in our methods, also replacing the \lr layer by a deeper classification \mlp, but there were no improvements.}  

\smallskip
\noindent \textbf{\darnn:} In a variant of \arnn, called \darnn (direct attention), the input to the first layer of the attention mechanism is the embedding $x_t$ of word $w_t$, rather than $h_t$ (cf.\ Eq.~\ref{eq:arnn}; $W^{(1, x)}\in \R^{r \times d}$):
\begin{equation}
a_t^{(1)} = \relu(W^{(1, x)} x_t + b^{(1)}) \label{eq:darnn}
\end{equation}
Intuitively, the attention of \arnn considers each word embedding $x_t$ in its (left) context, modelled by $h_t$, whereas the attention of \darnn considers directly $x_t$ without its context, but $h_{\textit{sum}}$ is still the weighted sum of the hidden states (Eq.~\ref{eq:hsum}).

\smallskip
\noindent \textbf{\eqrnn:} In another variant of \arnn, called \eqrnn, we assign equal attention to all the hidden states. The feature vector of the \lr layer is now the average $h_{\textit{sum}} =\frac{1}{k}\sum_{t=1}^{k} h_{t}$ (cf.\ Eq.~\ref{eq:hsum}).

\smallskip
\noindent \textbf{\danornn:} For ablation testing, we also experiment with a variant, called \danornn, that does not use the hidden states of the \rnn. The input to the attention mechanism is now directly the embedding $x_t$ instead of $h_t$ (as in \darnn, Eq.~\ref{eq:darnn}), and $h_{\textit{sum}}$ is the weighted average (centroid) of word embeddings $h_{\textit{sum}} =\sum_{t=1}^{k} a_{t} x_{t}$ (cf.\ Eq.~\ref{eq:hsum}).\footnote{We also tried \textit{tf-idf} scores in the $h_{\textit{sum}}$ of \danornn, instead of attention scores, but preliminary results were poor.}

\smallskip
\noindent \textbf{\nornn:} For further ablation, we also experiment with \nornn, which uses neither the \rnn nor the attention mechanism. The feature vector of the \lr layer is now simply the average of word embeddings $h_{\textit{sum}} =\frac{1}{k}\sum_{t=1}^{k} x_{t}$ (cf.\ Eq.~\ref{eq:hsum}).

\smallskip
We set $l=4, d = 300, m = r = 128$, having tuned the hyper-parameters of \rnn and \arnn on the same 2\% held-out training comments used to tune \Detox; \darnn, \eqrnn, \danornn, and \nornn use the same hyper-parameter values as \arnn, to make their results more directly comparable and save time. 
We use Glorot initialization \cite{Glorot2010}, cross-entropy loss, and Adam \cite{Kingma2015}.\footnote{We used Keras (\url{http://keras.io/}) with the TensorFlow back-end (\url{http://www.tensorflow.org/}).} Early stopping evaluates on the same held-out subsets. For Gazzetta, word embeddings are initialized to the \WordVec embeddings we provide (Section~\ref{sec:Gazzetta}). For the Wikipedia datasets, they are initialized to \glove embeddings \cite{Pennington2014}.\footnote{See \url{https://nlp.stanford.edu/projects/glove/}. We use `Common Crawl' (840B tokens).} In both cases, the embeddings are updated during backpropagation. Out of vocabulary (\oov) words, meaning words not encountered in the training set and/or words we have no initial embeddings for, are mapped (during training and testing) to a single randomly initialized embedding, which is also updated during training.\footnote{For Gazzetta, words encountered only once in the training set (\GazzettaTrain or \GazzettaTrainS) are also treated as \oov.}


\subsection{CNN} \label{sec:cnn}

We also compare against a vanilla \cnn operating on word embeddings. We describe the \cnn only briefly, because it is very similar to that of of Kim \shortcite{Kim2014}; see also Goldberg \shortcite{Goldberg2016} for an introduction to \cnn{s}, and Zhang and Wallace \shortcite{Zhang2015}. 

For Wikipedia comments, we use a `narrow' convolution layer, with kernels sliding (stride 1) over (entire) embeddings of word $n$-grams of sizes $n = 1, \dots, 4$. We use 300 kernels for each $n$ value, a total of 1,200 kernels. The outputs of each kernel, obtained by applying the kernel to the different $n$-grams of a comment $c$, are then max-pooled, leading to a single output per kernel. The resulting feature vector (1,200 max-pooled outputs) goes through a dropout layer \cite{Hinton2012} ($p=0.5$), and then to an \lr layer, which provides $P_{\cnn}(\textit{reject} | c)$. For Gazzetta, the \cnn is the same, except that $n = 1, \dots, 5$, leading to 1,500 features per comment. All hyper-parameters were tuned on the 2\% held-out training comments used to tune the other methods. Again, we use 300-dimensional word embeddings, which are now randomly initialized, since tuning indicated this was better than initializing to pre-trained embeddings. \oov words are treated as in the \rnn-based methods. All embeddings are updated.
Early stopping evaluates on the held-out subsets. Again, we use Glorot initialization, cross-entropy loss,  and Adam.\footnote{We implemented the \cnn directly in TensorFlow.} 


\subsection{LIST baseline} \label{sec:modlist}

A baseline, called \modlist, collects every word $w$ that occurs in more than 10 (for \DetoxAttTrain, \DetoxToxTrain, \GazzettaTrainS) or 100 comments (for \GazzettaTrain) in the training set, along with the precision of $w$, i.e., the ratio of rejected training comments containing $w$ divided by the total number of training comments containing $w$. The resulting lists contain 10,423, 11,360, 16,864, and 21,940 word types, when using \DetoxAttTrain, \DetoxToxTrain, \GazzettaTrainS, \GazzettaTrain, respectively.  For a comment $c$, $P_{\modlist}(\textit{reject} | c)$ is the maximum precision of all the words in $c$.


\subsection{Tuning thresholds} \label{sec:thresholds}

\begin{figure}
	\centering
	\includegraphics[width=0.85\columnwidth]{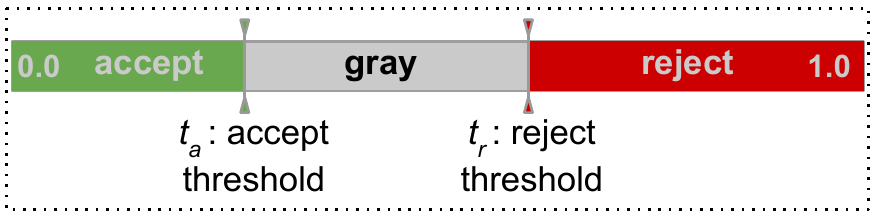}
	\caption{Illustration of threshold tuning.}
	\label{fig:tuning}
\end{figure}

All methods produce a $p = P(\textit{reject} | c)$ per comment $c$. In semi-automatic moderation (Fig.~\ref{fig:architecture}), a comment is directly rejected if its $p$ is above a rejection threshold $t_r$, it is directly accepted if $p$ is below an acceptance threshold $t_a$, and it is shown to a moderator if $t_a \leq p \leq t_r$ (gray zone of Fig.~\ref{fig:tuning}). 

In our experience, moderators (or their employers) can easily specify the approximate percentage of comments they can afford to check manually (e.g., 20\% daily) or, equivalently, the approximate percentage of comments the system should handle automatically. We call \emph{coverage} the latter percentage; hence, $1 - \textit{coverage}$ is the approximate percentage of comments to be checked manually. 
By contrast, moderators are baffled when asked to tune $t_r$ and $t_a$ directly. Consequently, we ask them to specify the approximate desired coverage. We then sort the comments of the development set (\GazzettaDev, \DetoxAttDev, 
\DetoxToxDev) by $p$, and slide $t_a$ from $0.0$ to $1.0$ (Fig.~\ref{fig:tuning}). For each $t_a$ value, we set $t_r$ to the value that leaves a $1 - \textit{coverage}$ percentage of development comments in the gray zone ($t_a \leq p \leq t_r$). We then select the $t_a$ (and $t_r$) that maximizes the weighted harmonic mean 
$F_{\beta}(P_{\textit{reject}}, P_{\textit{accept}})$ on the development set:
\[
F_{\beta}(P_{\textit{reject}}, P_{\textit{accept}}) = 
\frac{(1 + {\beta}^2)\cdot P_{\textit{reject}} \cdot P_{\textit{accept}}}
{{\beta}^2 \cdot P_{\textit{reject}} + P_{\textit{accept}}}
\]
where $P_{\textit{reject}}$ is the \emph{rejection precision} (correctly rejected comments divided by rejected comments) and $P_{\textit{accept}}$ is the \emph{acceptance precision} (correctly accepted divided by accepted). Intuitively, coverage sets the width of the gray zone, whereas $P_{\textit{reject}}$ and $P_{\textit{accept}}$ show how certain we can be that the red (reject) and green (accept) zones are free of misclassified comments. We set $\beta = 2$, emphasizing $P_{\textit{accept}}$, because moderators are more worried about wrongly accepting abusive comments than wrongly rejecting non-abusive ones.\footnote{More precisely, when computing $F_{\beta}$, we reorder the development comments by time posted, and split them into batches of 100. For each $t_a$ (and $t_r$) value, we compute $F_{\beta}$ per batch and macro-average across batches. The resulting thresholds lead to $F_{\beta}$  scores that are more stable over time.} The selected $t_a, t_r$ (tuned on development data) are then used in experiments on test data. In fully automatic moderation, $\textit{coverage} = 100\%$ and $t_a = t_r$; otherwise, threshold tuning is identical.  


\section{Experimental results} \label{sec:results}

Following Wulczyn et al.\ \shortcite{Wulczyn2016}, we report in Tables~\ref{tab:gazzetta_eval_scores}--\ref{tab:wikipedia_eval_scores} \auc scores (area under \textsc{roc} curve), along with Spearman correlations between system-generated probabilities $P(\textit{accept}|c)$ and human probabilistic gold labels (Section~\ref{sec:wikipedia}) when probabilistic gold labels are available.\footnote{When computing \auc, the gold label is the majority label of the annotators. When computing Spearman, the gold label is probabilistic (\% of annotators that accepted the comment). The decisions of the systems are always probabilistic.} 

\begin{table}
\centering
{\small
\begin{tabular}{|c|c|c|c|c|c|}
\hline
\multicolumn{6}{|c|}{\textbf{Training dataset: \GazzettaTrainS}}\\\hline
\multirow{2}{*}{System} & \GazzettaDev & \GazzettaTest & \GazzettaClean & \multicolumn{2}{|c|}{\GazzettaCleanR}\\\cline{2-6}
                        & \auc         & \auc          & \auc           & \auc  & Spearman\\\hline
\rnn                    & 75.75        & 75.10         & 74.40          & 80.27 & 51.89 \\\hline
\arnn                   & \textbf{76.19}        & \textbf{76.15}         & \textbf{75.83}          & \textbf{80.41} & \textbf{52.51} \\\hline
\darnn                  & 75.96        & 75.90         & 74.25          & 80.05 & 52.49 \\\hline
\eqrnn                  & 74.31        & 74.01         & 73.28          & 77.73 & 45.77 \\\hline
\danornn                & 75.09        & 74.96         & 74.20          & 79.92 & 51.04 \\\hline
\nornn                  & 73.93        & 73.82         & 73.80          & 78.45 & 48.14\\\hline
\cnn                    & 70.97        & 71.34         & 70.88          & 76.03 & 42.88 \\\hline
\Detox                  & 72.50        & 72.06         & 71.59          & 75.67 & 43.80 \\\hline
\modlist                & 61.47        & 61.59         & 61.26          & 64.19 & 24.33 \\\hline\hline
\multicolumn{6}{|c|}{\textbf{Training dataset: \GazzettaTrain}}\\\hline
\multirow{2}{*}{System} & \GazzettaDev & \GazzettaTest & \GazzettaClean & \multicolumn{2}{|c|}{\GazzettaCleanR}\\\cline{2-6}
                        & \auc         & \auc          & \auc           & \auc  & Spearman\\\hline
\rnn                    & 79.50        & 79.41         & 79.23          & 84.17 & 59.31 \\\hline
\arnn                   & \textbf{79.64}        & \textbf{79.58}         & \textbf{79.67}          & \textbf{84.69} & \textbf{60.87} \\\hline
\darnn                  & 79.60        & 79.56         & 79.38          & 84.40 & 60.83\\\hline
\eqrnn                  & 77.45        & 77.76         & 77.28          & 82.11 & 55.01 \\\hline
\danornn                & 78.73        & 78.64         & 78.62          & 83.53 & 57.82 \\\hline
\nornn                  & 76.76 & 76.85 & 76.30 & 82.38 & 53.28\\\hline
\cnn                    & 77.57        & 77.35         & 78.16          & 83.98 & 55.90 \\\hline
\Detox                  & --           & --            & --             & --    & --    \\\hline
\modlist                & 67.04        & 67.06         & 66.17          & 69.51 & 33.61 \\\hline
\end{tabular}
}
\caption{Results on Gazzetta comments.}
\label{tab:gazzetta_eval_scores}
\end{table}

A first observation is that increasing the size of the Gazzetta training set (\GazzettaTrainS to \GazzettaTrain, Table~\ref{tab:gazzetta_eval_scores}) significantly improves the performance of all methods; we do not report \Detox results for \GazzettaTrain, because its implementation could not handle the size of \GazzettaTrain. Tables~\ref{tab:gazzetta_eval_scores}--\ref{tab:wikipedia_eval_scores} also show that \rnn is always better than \cnn and \Detox; there is no clear winner between \cnn and \Detox. Furthermore, \arnn is always better than \rnn on Gazzetta comments (Table~\ref{tab:gazzetta_eval_scores}), but not always on Wikipedia comments (Table~\ref{tab:wikipedia_eval_scores}). 
Another observation is that \darnn is always worse than \arnn (Tables~\ref{tab:gazzetta_eval_scores}--\ref{tab:wikipedia_eval_scores}), confirming that the hidden states of the \rnn 
are a better input to the attention mechanism than word embeddings. 
The performance of \darnn deteriorates further when equal attention is assigned to the hidden states (\eqrnn), when the weighted sum of hidden states ($h_{\textit{sum}}$) is replaced by the weighted sum of word embeddings (\danornn), or both (\nornn).  Also, \danornn outperforms \nornn, indicating that the attention mechanism improves the performance of simply averaging word embeddings. The Wikipedia subsets are easier (all methods perform better on Wikipedia subsets, compared to Gazzetta).

\begin{table}
\centering
{\small
\begin{tabular}{|c|c|c|c|c|}
\hline
\multicolumn{5}{|c|}{\textbf{Training dataset: \DetoxAttTrain}}\\\hline
\multirow{2}{*}{System} & \multicolumn{2}{|c|}{\DetoxAttDev} & \multicolumn{2}{|c|}{\DetoxAttTest}\\\cline{2-5}
                        & \auc  & Spearman & \auc  & Spearman\\\hline
\rnn                    & 97.39 & \textbf{71.92}    & \textbf{97.71} & \textbf{72.79}   \\\hline
\arnn                   & \textbf{97.46} & 71.59    & 97.68 & 72.32   \\\hline
\darnn                  & 97.02 & 71.49 & 97.31 & 72.11 \\\hline
\eqrnn                  & 92.66 & 60.77 & 92.85 & 60.16 \\\hline
\danornn                & 96.73 & 70.13 & 97.06 & 71.08 \\\hline
\nornn                  & 92.30 & 57.21 & 92.81 & 56.33 \\\hline
\cnn                    & 96.91 & 70.06 & 97.07 & 70.21 \\\hline
\Detox                  & 96.26 & 67.75    & 96.71 & 68.09   \\\hline
\modlist                & 93.05 & 55.39    & 92.91 & 54.55   \\\hline
\hline
\multicolumn{5}{|c|}{\textbf{Training dataset: \DetoxToxTrain}}\\\hline
\multirow{2}{*}{System} & \multicolumn{2}{|c|}{\DetoxToxDev} & \multicolumn{2}{|c|}{\DetoxToxTest}\\\cline{2-5}
& \auc  & Spearman & \auc  & Spearman\\\hline
\rnn                    & 98.20 & 68.84 & \textbf{98.42} & 68.89\\\hline
\arnn                   & \textbf{98.22} & \textbf{68.95} & 98.38 & \textbf{68.90}\\\hline
\darnn                  & 98.05 & 68.59 & 98.28 & 68.55 \\\hline
\eqrnn                  & 94.72 & 55.48 & 95.04 & 55.86 \\\hline
\danornn                & 97.83 & 67.86 & 97.94 & 67.74\\\hline
\nornn                  & 94.31 & 53.35 & 94.61 & 52.93\\\hline
\cnn                    & 97.76 & 65.50 & 97.86 & 65.56 \\\hline
\Detox                  & 97.16 & 63.57 & 97.13 & 63.24 \\\hline
\modlist                & 93.96 & 51.35 & 93.95 & 51.18 \\\hline
\end{tabular}
}
\caption{Results on Wikipedia comments.
}
\label{tab:wikipedia_eval_scores}
\end{table}

\begin{figure}
	\centering
	\includegraphics[width=\columnwidth]{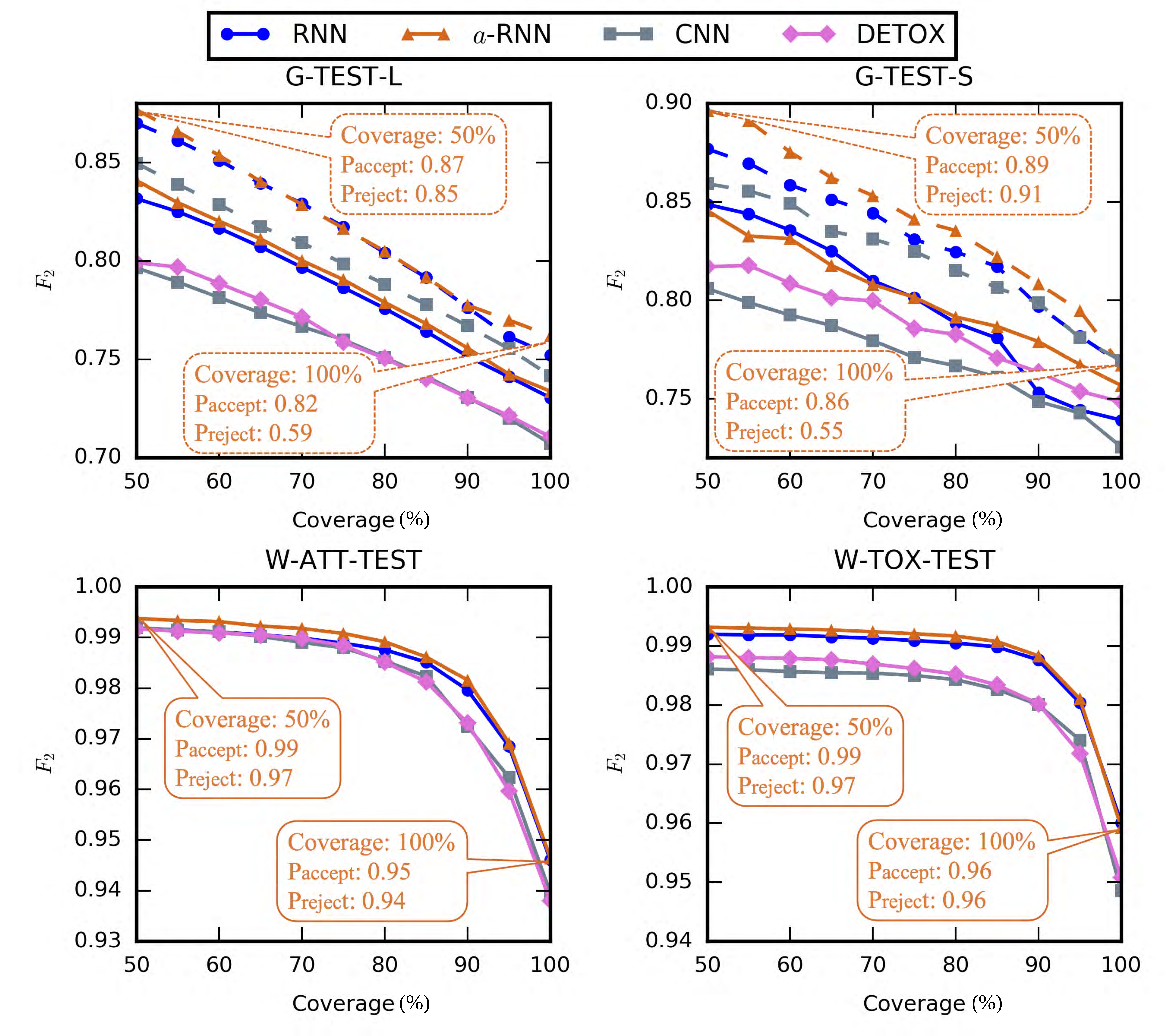}
	\caption{$F_2$ scores for varying coverage. Dashed lines were obtained using a larger training set.}
	\label{fig:f2_per_coverage}
\end{figure}

Figure~\ref{fig:f2_per_coverage} shows $F_2(P_{\textit{reject}}, P_{\textit{accept}})$ on \GazzettaTest, \GazzettaClean, \DetoxAttTest, \DetoxToxTest, when $t_a, t_r$ are tuned on the corresponding development tests for varying coverage. For the Gazzetta datasets, we show results training on \GazzettaTrainS (solid lines) and \GazzettaTrain (dashed). The differences between \rnn and \arnn are again small, but it is now easier to see that \arnn is overall better. Again, \arnn and \rnn are better than \cnn and \Detox, and the results improve with a larger training set (dashed). On \DetoxAttTest and \DetoxToxTest, \arnn obtains $P_{\textit{accept}}, P_{\textit{reject}} \geq 0.94$ for all coverages (Fig.~\ref{fig:f2_per_coverage}, call-outs). On the more difficult Gazzetta datasets, \arnn still obtains $P_{\textit{accept}}, P_{\textit{reject}} \geq 0.85$ when tuned for 50\% coverage. When tuned for 100\% coverage, comments for which the system is uncertain (gray zone) cannot be avoided and there are inevitably more misclassifications; the use of $F_2$ during threshold tuning places more emphasis on avoiding wrongly accepted comments, leading to high $P_{\textit{accept}}$ ($\geq 0.82$), at the expense of wrongly rejected comments, i.e., sacrificing $P_{\textit{reject}}$ ($\geq 0.56$). On the re-moderated \GazzettaCleanR (similar diagrams, not shown), $P_{\textit{accept}}, P_{\textit{reject}}$ become 0.96, 0.88 for coverage 50\%, and 0.92, 0.48 for coverage 100\%. 

\section{Related work}

Napoles et al.\ \shortcite{Napoles2017a} developed an annotation scheme for online conversations, with 6 dimensions for comments (e.g., sentiment, tone, off-topic) and 3 dimensions for threads. The scheme was used to label a dataset, called \textsc{ynacc}, of 9.2K comments (2.4K threads) from Yahoo News and 16.6K comments (1K threads) from the Internet Argument Corpus \cite{Walker2012,Abbott2016}. Abusive comments were filtered out, hence \textsc{ynacc} cannot be used for our purposes, but it may be possible to extend the annotation scheme for abusive comments, to predict more fine-grained labels, instead of `accept' or `reject'. Napoles et al.\ also reported that up/down votes, a form of social filtering, are inappropriate proxies for comment and thread quality. Lee et al.\ \shortcite{Lee2014} discuss social filtering in detail and propose features (e.g., thread depth, no.\ of revisiting users)  to assess the quality of a thread without processing the texts of its comments. Diakopoulos \shortcite{Diakopoulos2015} discusses how editors select high quality comments. 

In further work, Napoles et al.\ \shortcite{Napoles2017b} aimed to identify high quality threads. Their best method converts each comment to a comment embedding using \textsc{doc2vec} \cite{Le2014}. An ensemble of Conditional Random Fields (\crf{s}) \cite{Lafferty2001} assigns labels (from their annotation scheme, e.g., for sentiment, off-topic) to the comments of each thread, viewing each thread as a sequence of \textsc{doc2vec} embeddings. The decisions of the \crf{s} are then used to convert each thread to a feature vector (total count and mean marginal probability of each label in the thread), which is passed on to an \lr classifier. Further improvements were observed when additional features were added, \bow counts and \pos $n$-grams being the most important ones. Napoles et al.\ \shortcite{Napoles2017b} also experimented with a \cnn, similar to that of Section~\ref{sec:cnn}, which was not however a top-performer, presumably because of the small size of the training set (2.1K \textsc{ynacc} threads). 

Djuric et al.\ \shortcite{Djuric2015} experimented with 952K manually moderated comments from Yahoo Finance, but their dataset is not publicly available. They convert each comment to a \textsc{doc2vec} embedding, which is fed to an \lr classifier. Nobata et al.\ \shortcite{Nobata2016} experimented with approx.\ 3.3M manually moderated comments from Yahoo Finance and News; their data are also not available.\footnote{According to Nobata et al., their clean test dataset (2K comments) would be made available, but it is currently not.} They used Vowpal Wabbit\footnote{See \url{http://hunch.net/~vw/}.} with character $n$-grams ($n = 3,\dots, 5$) and word $n$-grams ($n = 1, 2$), hand-crafted features (e.g., comment length, number of capitalized or black-listed words), features based on dependency trees, averages of \WordVec embeddings, and \textsc{doc2vec}-like embeddings. Character $n$-grams were the best, on their own outperforming Djuric et al.\ \shortcite{Djuric2015}. The best results, however, were obtained using all features. By contrast, we use no hand-crafted features and parsers, making our methods easily portable to other domains and languages. 

Wulczyn et al.\ \shortcite{Wulczyn2016} experimented with character and word $n$-grams, based on the findings of Nobata et al.\ \shortcite{Nobata2016}. We included their dataset and moderation system (\Detox) in our experiments. Wulczyn et al.\ also used \Detox (trained on \DetoxAttTrain) as a proxy (instead of human annotators) to automatically classify 63M Wikipedia comments, which were then used to study the problem of personal attacks (e.g., the effect of allowing anonymous comments, how often personal attacks were followed by moderation actions). Our methods could replace \Detox in studies of this kind, since they perform better.

Waseem et al.\ \shortcite{Waseem2016} used approx.\ 17K tweets annotated for hate speech. Their best method was an \lr classifier with character $n$-grams ($n = 1, \dots, 4$) and a gender feature. Badjatiya et al.\ \shortcite{Badjatiya2017} experimented with the same dataset using \lr, \svm{s} \cite{Cortes1995}, Random Forests \cite{Ho1995}, Gradient Boosted Decision Trees (\gbdt) \cite{Friedman2002}, \cnn (similar to that of Section~\ref{sec:cnn}), \lstm \cite{Greff2015}, FastText \cite{Joulin2017}. They also considered alternative feature sets: character $n$-grams, \textit{tf-idf} vectors, word embeddings, averaged word embeddings. Their best results were obained using \gbdt with averaged word embeddings learned by the \lstm, starting from random embeddings.

Warner and Hirschberg \shortcite{Warner2012} aimed to detect anti-semitic speech, experimenting with 9K paragraphs and a linear \svm. Their features consider windows of up to 5 tokens, the tokens of each window, their order, \textsc{pos} tags, Brown clusters etc., following Yarowsky \shortcite{Yarowsky1994}. 

Cheng et al.\ \shortcite{Cheng2015} predict which users would be banned from on-line communities. Their best system uses a Random Forest or \lr classifier, with features examining readability, activity (e.g., number of posts daily),
community and moderator reactions (e.g., up-votes, number of deleted posts).  

Lukin and Walker \shortcite{Lukin2013} experimented with 5.5K utterances from the Internet Argument Corpus \cite{Walker2012,Abbott2016} annotated with nastiness scores, and 9.9K utterances from the same corpus annotated for sarcasm.\footnote{For sarcasm, 
see Davidov et al.\ \shortcite{Davidov2010}, Gonzalez-Ibanez et al.\ \shortcite{GonzlezIbez2011}, Joshi et al.\ \shortcite{Joshi2015}, Oraby et al.\ \shortcite{Oraby2016}.} In a bootstrapping manner, they manually identified cue words and phrases (indicative of nastiness or sarcasm), used the cue words to obtain training comments, and extracted patterns from the training comments. Xiang et al.\ \shortcite{Xiang2012} also employed bootstrapping to identify users whose tweets frequently or never contain profane words, and collected 381M tweets from the two user types. They  trained decision tree, Random Forest, or \lr classifiers to distinguish between tweets from the two user types, testing on 4K tweets manually labeled as containing profanity or not. The classifiers used topical features, obtained via \textsc{lda} \cite{Blei2003}, and a feature indicating the presence of at least one of approx.\ 330 known profane words.

Sood et al.\ \shortcite{Sood2012,Sood2012b} experimented with 6.5K comments from Yahoo Buzz, moderated via crowdsourcing. They showed that a linear \svm, representing each comment as a bag of word bigrams and stems, performs better than word lists. Their best results were obtained by combining the \svm with a word list and edit distance.

Yin et al.\ \shortcite{Yin2009} used posts from chat rooms and discussion fora ($<$15K posts in total) to train an \svm to detect online harassment. They used \textsc{tf-idf}, sentiment, and context features (e.g., similarity to other posts in a thread).\footnote{Sentiment features have been used by several methods, but sentiment analysis \cite{Pang2008,Liu2015b} is typically not directly concerned with abusive content.} Our methods might also benefit by considering threads, rather than individual comments. 
Yin et al.\ point out that unlike other abusive content, spam in comments or discussion fora \cite{Michne2005, Niu2007} is off-topic and serves a commercial purpose. Spam is unlikely in Wikipedia discussions and extremely rare so far in Gazzetta comments.

Mihaylov and Nakov \shortcite{Mihaylov2016} identify comments posted by opinion manipulation trolls. Dinakar et al.\ \shortcite{Dinakar2011} and Dadvar et al.\ \shortcite{Dadvar2013} detect cyberbullying. Chandrinos et al.\ \shortcite{Chandrinos2000} detect pornographic web pages, using a Naive Bayes classifier with text and image features.
Spertus \shortcite{Spertus1997} flag flame messages in Web feedback forms, using decision trees and hand-crafted features. A Kaggle dataset for insult detection is also available.\footnote{See \url{http://www.kaggle.com/}, data description of the competition `Detecting Insults in Social Commentary'.} It contains 6.6K comments
(3,947 train, 2,647 test) labeled as insults or not. However, abusive comments that do not directly insult other participants of the same discussion are not classified as insults, even if they contain profanity, hate speech, insults to third persons etc.


\section{Conclusions} \label{sec:conc}

We experimented with a new publicly available dataset of 1.6M moderated user comments from a Greek sports news portal and two existing datasets of English Wikipedia talk page comments. We showed that a \gru \rnn operating on word embeddings outperforms the previous state of the art, which used an \lr or \mlp classifier with character or word $n$-gram features. It also outperforms a vanilla \cnn  operating on word embeddings, and a baseline that uses an automatically constructed word list with precision scores. A novel, deep, classification-specific attention mechanism improves further the overall results of the \rnn. The attention mechanism also improves the results of a simpler method that averages word embeddings. We considered both fully automatic and semi-automatic moderation, along with threshold tuning and evaluation measures for both.  

We plan to consider user-specific information (e.g., ratio of comments rejected in the past) and thread statistics (e.g., thread depth, number of revisiting users) \cite{Dadvar2013,Lee2014,Cheng2015,Waseem2016}. We also plan to explore character-level \rnn{s} or \cnn{s} \cite{Zhang2015b}, for example to produce embeddings of unknown or obfuscated words from characters \cite{Santos2014,Ling2015}. We are also exploring how the attention scores of \arnn can be used to highlight `suspicious' words or phrases when showing gray comments to moderators.


\section*{Acknowledgments}

This work was funded by Google's Digital News Initiative (project \textsc{ml2p}, contract 362826).\footnote{See \url{https://digitalnewsinitiative.com/}.} We are grateful to Gazzetta for the data they provided. We also thank Gazzetta's moderators for their feedback, insights, and advice.

\bibliography{acl_abusive_2017}
\bibliographystyle{acl_natbib}
\end{document}